\documentclass[a4paper,conference]{IEEEtran}

\ifCLASSINFOpdf
\else
\fi

\usepackage{graphicx}
\usepackage{comment}
\usepackage{amsmath,amssymb} 
\usepackage{color}
\usepackage[graphicx]{realboxes}
\usepackage{xcolor}
\usepackage{array}
\usepackage{adjustbox} 
\usepackage{pbox}
\usepackage{float}   
\usepackage{cite}
\usepackage{amsmath,amssymb,amsfonts}
\usepackage{algorithmic}
\usepackage{indentfirst}
\usepackage{sfmath}
\usepackage{mathdots}
\usepackage{yhmath}
\usepackage{color}

\usepackage{multirow}

\DeclareMathOperator{\arctantwo}{\mathnormal{arctan2}}

\begin{document}
%
\title{P2D: a self-supervised method for depth estimation from polarimetry}


\author{Marc Blanchon$^{1\dag}$, D\'esir\'e Sidib\'e$^{2}$, Olivier Morel$^{1}$, Ralph Seulin$^{1}$, Daniel Braun$^{1}$ and Fabrice Meriaudeau$^{1}$\\
$^{1}$ERL VIBOT CNRS 6000, ImViA, Universit\'e Bourgogne Franche-Comt\'e, 71710, Le Creusot, France\\
$^{2}$IBISC, Univ Evry, Universit\'e Paris-Saclay, 91025, Evry, France\\
$^{\dag}$Email: marc.blanchon@u-bourgogne.fr\\
}


%


\maketitle

\begin{abstract}
	Monocular depth estimation is a recurring subject in the field of computer vision. Its ability to describe scenes via a depth map while reducing the constraints related to the formulation of perspective geometry tends to favor its use.
	    However, despite the constant improvement of algorithms, most methods exploit only colorimetric information. Consequently, robustness to events to which the modality is not sensitive to, like specularity or transparency, is neglected. In response to this phenomenon, we propose using polarimetry as an input for a self-supervised monodepth network. 
        Therefore, we propose exploiting polarization cues to encourage accurate reconstruction of scenes. Furthermore, we include a term of polarimetric regularization to state-of-the-art method to take specific advantage of the data. Our method is evaluated both qualitatively and quantitatively demonstrating that the contribution of this new information as well as an enhanced loss function improves depth estimation results, especially for specular areas.
\end{abstract}


%
\IEEEpeerreviewmaketitle

	\section{Introduction}

	Depth estimation represents an important subject in the field of computer vision, and numerous methods were developed involving geometry and optimization processes.
At present, algorithms tend to be artificially lightened by the use of deep learning \cite{jiang2018self,Zhang_2019_ICCV,jiao2018look}. 
Moreover, a trend towards the relaxation of some constraints can be observed typically with the monodepth approaches \cite{Dijk_2019_ICCV,Hu_2019_ICCV,Chen_2019_CVPR}.
One of the common points of all these modern algorithms is they predominantly use a single dataset: KITTI \cite{Geiger2012CVPR,Menze2015CVPR}. This dataset is composed of a massive number of RGB images which greatly links the sensory system to the applications. 
As a result, the modality limitations are exported to the applications and some problematic aspects such as specular surfaces are neglected.
By following the logic of depth estimation via a monocular camera, it is possible to provide a new modality able to address the weaknesses of RGB data. In this perspective, polarimetry seems to represent an excellent candidate for depth estimation in urban areas. Indeed, by definition, polarimetric data are particularly able to cope with light phenomena. Moreover, the applications of polarimetry were previously highly specialized because of the cost of the sensors and the specificity of the very constraining acquisition. Nowadays, modern devices tend to both reduce their cost and facilitate dynamic acquisition via focal plane division sensors which induce increasingly popular uses
~\cite{berger2017depth,cui2017polarimetric,rastgoo2018attitude,nguyen20173d}. 
In addition, it has been established a deep learning network requires a large amount of diverse data to learn relevant representations. As follows, the scarcity of polarimetric datasets tends to encourage the use of monocular depth estimation methods via self-supervised deep learning.

	\begin{figure}[t]
		\centering
		\includegraphics[keepaspectratio,width=\linewidth]{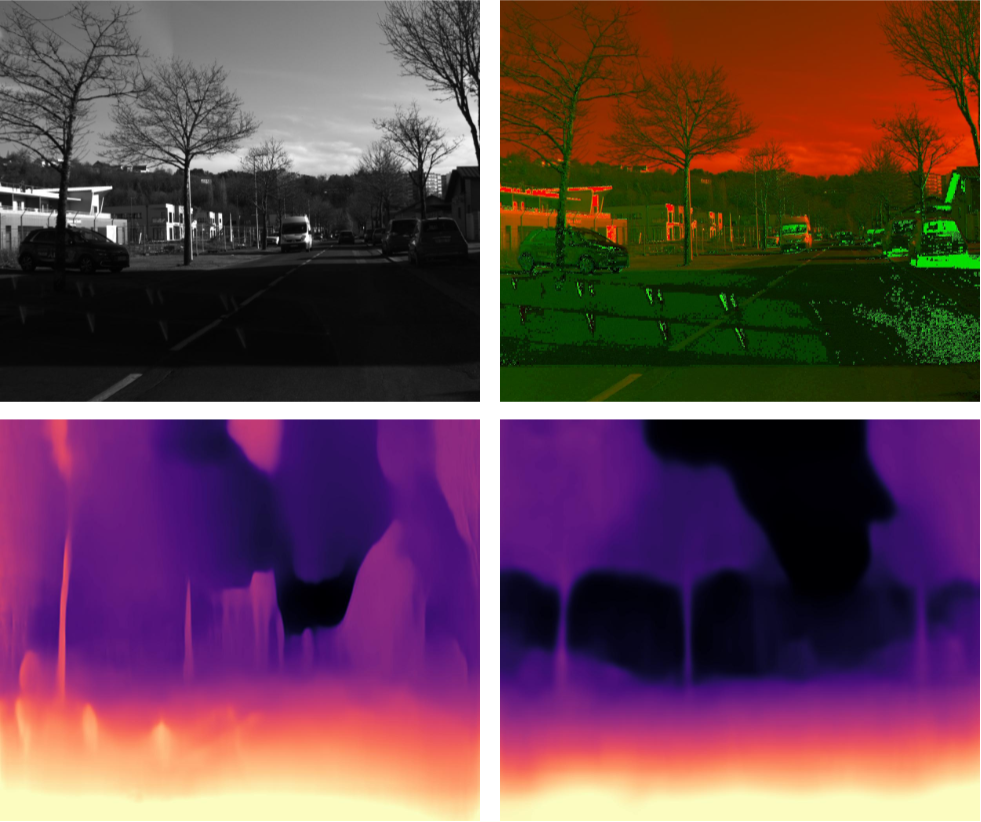}
\caption{Illustration of the proposed method compared to a state-of-the-art method~\cite{godard2019digging}. On the left are the input grayscale image and the corresponding disparity obtained by \cite{godard2019digging}. On the right are the polarimetric input and the  result of the proposed P2D.}
		\label{res1}
	\end{figure}
	
	\indent In this paper, we present P2D, a self-supervised monocular depth estimation network taking advantage of polarization cues. 
	As shown in Figure~\ref{res1}, we exploit the polarimetric information to make the estimation sensitive to specularity as illustrated by the plane reconstruction on the car windows. In addition, since the algorithms are already efficient for diffuse areas, we propose not to penalize these estimates but to improve their weaknesses.
	Since polarimetric data is scarce, we exploit the faculties of self-supervision to train on urban dataset acquired with polarization sensor for the occasion. 
	The particular issue we address is very challenging as it confronts the problems of specularity in urban scenes. 
 We propose an evaluation of a recent deep learning based technique and of our method on a benchmark allowing a common and fair comparison.
	
	\section{Related Works}
	This section will discuss some previous works related to the two main components of this contribution.
	
	\indent\textbf{Polarimetry and geometry.} 
Polarimetry is directly related to scene geometry. As light interacts with materials, this unique modality acquires the relationships between the two.
This property is unrestrictive since this type of data does not uniquely define specular areas but also diffuse areas. Therefore, the use of such modality could be profitable since it provides another complementary information to RGB. Moreover, when a reflection can be clearly defined by polarization, the RGB values tend to saturate or mislead algorithms.\\
Many algorithms take advantage of polarimetric information such as: Structure from Polarization \cite{rahmann2001reconstruction,morel2006active,cui2017polarimetric}, classification of urban scenes \cite{blin2020new} and pose estimation~\cite{shabayek2012vision,rastgoo2018attitude}.
Also, while using existing methods adapted to this modality, some contributions propose improved versions of scene geometry estimation. In \cite{berger2017depth}, an efficient stereovision method is proposed, reconstructing specular areas by minimizing a cost function inspired by \cite{Woodford2008GlobalPriors}. Other approaches induce concepts of multiple view geometry and tend to require a more constraining acquisition system as in \cite{cui2017polarimetric,yang2018polarimetric}.
Ultimately, some methods propose alternatives requiring either an increased knowledge of the scene via the refractive index \cite{kadambi2015polarized} or an improved acquisition system \cite{DBLP:journals/corr/abs-1903-12061,keller2019stereo}.
Despite an increasing diversity of reconstruction methods with polarization cues, the approaches are still ruled by Fresnel equations linking polarization, materials and normals. In addition, there is a tendency to use polarimetry either as an addition to other modalities or in complex multi-camera/image systems. \\
It is therefore clear that polarimetry remains a legitimate contender when it comes to taking advantage of scene geometry. Especially since most real-world acquisitions are uncontrolled and therefore scenes are prone to contain specular areas. It is also noteworthy that in an urban environment, the majority of priority areas are reflective (car, road, etc.).\\
In essence, polarimetric modality is capable of characterizing geometry by itself. This capacity is inherent to an in-depth understanding of the scene (materials, pose, light, etc.). In an unconstrained environment like urban areas, the lack of this information can cause most estimation methods to fail. Our approach proposes overcoming these constraints by using the abstraction capabilities of a deep learning network.
Consequently, our method does not require a deep understanding of the scene and is therefore usable in uncontrolled urban areas.\\

	\textbf{Monodepth.} There is an increased interest in the study of monocular depth estimation methods, since they offer a more flexible acquisition system as opposed to stereo systems. Most of the works are based on deep learning and optimize specific cost functions. We propose outlining some key methods that either express the basics for the domain, or similarly to our approach, impose unique constraints to the cost function refining the estimated maps.\\
Recent approaches alternate between training with a stereovision pair~\cite{zhan2018unsupervised,yang2018every,garg2016unsupervised} and a monocular one \cite{godard2019digging,godard2017unsupervised,ranjan2019competitive}.
A frequent feature of these algorithms is they are principally based on a cost function including two terms: reconstruction and smoothing.
Most of the current improvements propose introducing new constraints to refine the results.
\cite{watson2019self} shows the contribution of a stereo prior cue allows an improvement in accuracy. Motion estimation has also proved its value when refining a depth map \cite{brickwedde2019mono}.
Furthermore, many other features like semantics \cite{chen2019towards}, structured light pattern \cite{riegler2019connecting} or continuous Conditional Random Fields \cite{xu2017multi} seem to improve the estimation on an ad-hoc basis.
Ultimately, the majority of the works are based on photometric features and does not transfer the achievements in non-visible image spectrum.\\
Our self-supervised approach introduces a possibility to append other physics-to-geometry based constraints. To be specific, we address the problem of adding polarization cues in the reconstruction cost function. Including an understanding of more complex scenes therefore reports an improving result especially on specular surfaces.

	\section{Method}
	The proposed approach is primarily based on the works of Godard et al. \cite{godard2019digging} and Berger et al. \cite{berger2017depth}. In this section, we will describe our approach constraining the loss proposed in~\cite{godard2019digging} by considering polarization cues. We will subsequently discuss the methods used to train the network as well as the constraints related to the modality used.\\
	
	\textbf{Network and initial loss.} 
    	Our approach is based on an architecture similar to the one proposed in \cite{godard2019digging} since the formulation of the loss implies geometric constraints, partly explaining the accuracy of the inference.
Moreover, the key strengths of such network remain the self-supervision as well as the subdivision of the task into two parts, which makes the procedure highly adaptable.
The method is self-supervised thanks to the loss which by definition does not require any ground truth. The comparison metric used in back propagation is evaluated through network output and input. The geometrically consistent loss is expressed according to a photometric reprojection error and a smoothness term, which will be discussed farther below.\\
	
	\textbf{Basics of polarimety.} Polarimetry represents a modality sensitive to the interaction of light with the environment, thus allowing the characterization of both specular and diffuse areas of a scene.

Polarimetric information is conventionally indexed by Stokes parameters $S$ depending on the polarizers intensities oriented at different angles $P_{\{0,45,90,135\}}$:
	
	\begin{equation}
	S = \begin{pmatrix}s_0\\s_1\\s_2\\s_3\end{pmatrix} = \begin{pmatrix}P_0 + P_{90}\\ P_0 - P_{90}\\ P_{45} - P_{135} \\ 0\end{pmatrix}, 
	\end{equation}
	with $s_3$ remaining null since the circular polarization is not acquired. From these Stokes parameters, one can derive three informative images characteristic of polarimetry \cite{collett2005field}, polarization angle $\alpha$, degree of polarization $\rho$ and intensity $\iota$:
	
	\begin{equation}\label{polaparam}
	\begin{cases}  \rho = \sqrt{(\frac{s_1}{s_0})^2 + (\frac{s_2}{s_0})^2} \\[8pt] 
	
	\alpha = \frac{1}{2}\arctantwo(s_1, s_2) \\[8pt]
	
	\iota = \frac{P_0 + P_{45} + P_{90} + P_{135}}{2}
	
	\end{cases}.
	\end{equation}

	\begin{figure}[t]
		\centering
		\vspace{0.1cm}
		\includegraphics[keepaspectratio,width=.8\linewidth]{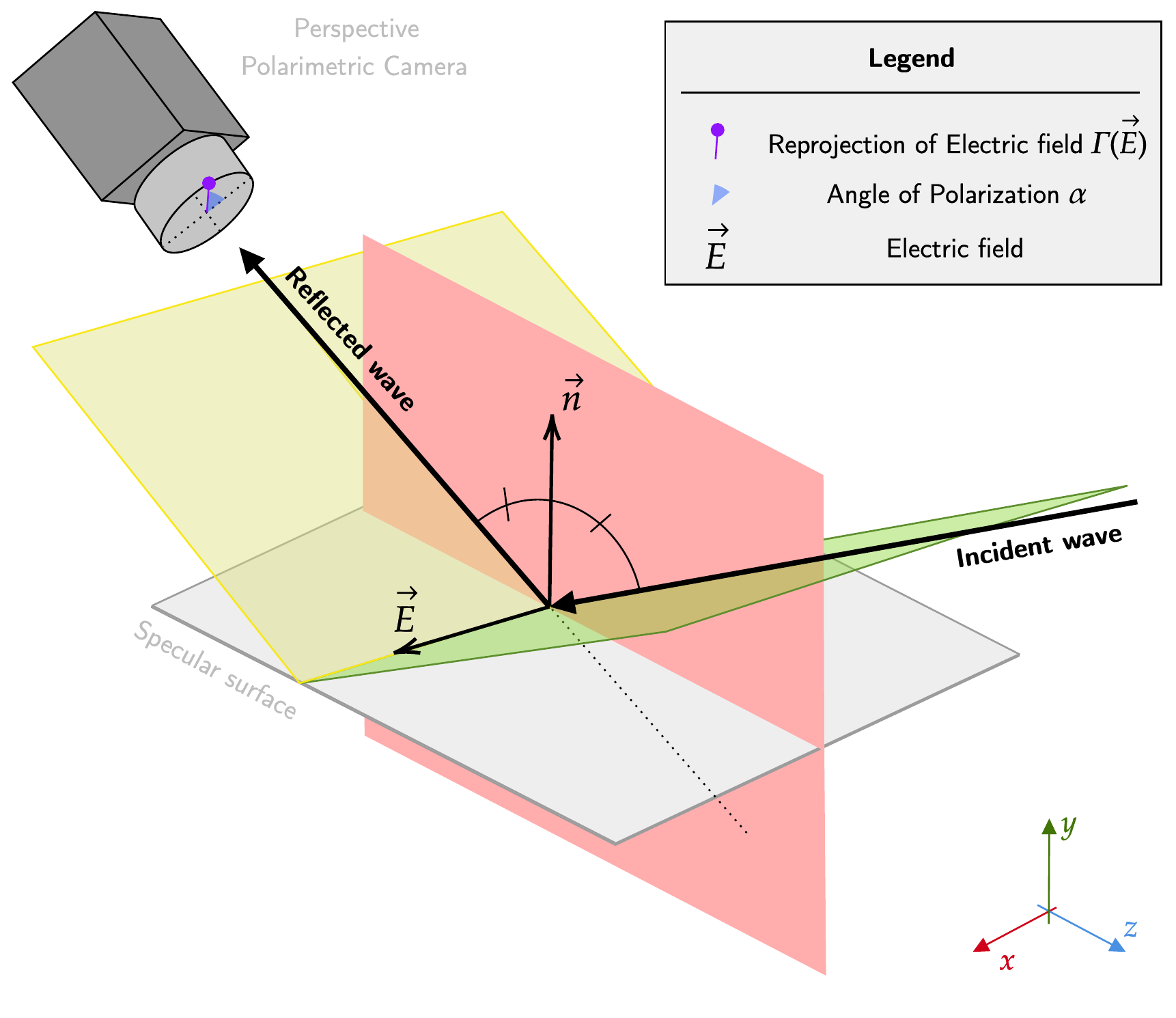}
		\caption{Illustration of imbrication of polarimetry peculiar and perspective geometry. Visual representation of angle of polarization measurement.}
		\label{fig:found}
	\end{figure}

	The intensity is then scalar, while the degree of polarization is analogous to the strength of the polarization.
	As shown in Figure~\ref{fig:found}, it is equally possible to geometrically represent some components of polarimetric information mixing peculiar and perspective geometry. The polarization angle $\alpha$ results from the angle between the reference and the direction of the electric field projection.\\

	\textbf{Prior polarimetric reconstruction error.} The paper by Berger et al. \cite{berger2017depth} proposes an approach to minimize an error specific to polarimetric induced geometry. Drawing on the terms provided by Woodford et al. \cite{Woodford2008GlobalPriors}, the method consists in including a minimizable expression compelling a normal/polarization angle consistency.
It is consequently shown that constraining a cost function involving a polarimetry-specific geometry is valid. Furthermore, this minimization approach is operable when optimizing a deep learning model since it depends on both the input and output of the processing pipeline and therefore could guarantees a self-supervision capability. Nevertheless, the acquisition setup as well as the problem formulation highly influence the error calculation. Indeed, \cite{berger2017depth} proposed an azimuth to acquired angle of polarization comparison. This approach is consistent under peculiar conditions implying restricted calibration of the camera or azimuth to angle of polarization specific link hypothesis. For this reason, our method proposes an alternative but similar approach allowing standard calibration and a generalized loss term releasing the constraints and allowing for easier use in real word applications. \\
	
	\textbf{Constraining the loss.} Fundamentally, the function to minimize includes a reprojection error term and a smoothing term. Our method P2D employs the photometric error proposed in~\cite{godard2019digging} since it has demonstrated its optimization and efficient convergence capabilities via deep learning. As a replacement for the edge aware first order smoothness, the second order derivative enhancement proposed by \cite{Woodford2008GlobalPriors} is used to encourage fine transitions counterbalancing the discontinuities induced by the polarization parameters.\\ 
	First, one penalizes the photometric reprojection error:
	
	\begin{equation}
	L_r = \min_{t^\prime} \: pe(I_t, I_{t^\prime \rightarrow t}),
	\end{equation}
	with $t^\prime \rightarrow t$ the pose transformation between two consecutive views and $pe$ the reconstruction error:
	
	\begin{equation}
	pe(I_a, I_b) = \frac{\beta}{2} (1- SSIM(I_a, I_b)) + (1-\beta) ||I_a - I_b||_1.
	\end{equation}

	To comply with the specifications of a minimizable function, the reprojection error comprises the weighted combination of structural dissimilarity (DSSIM) and L1 difference penalizing the deviation per pixel of the reprojection. As described in the original paper, $\beta = 0.85$ is used.

	In a second step, a smoothing term is used to encourage a precise estimation of the planes while taking into account the edges:
	
	\begin{equation}
	L_s = |\delta^2_xd_t^*| e^{-|\delta^2_x I_t|} + |\delta^2_yd_t^*| e^{-|\delta^2_y I_t|},
	\end{equation}
	where $d_t^* = d_t / \bar{d_t}$ is the mean-normalized inverse depth enforcing the depth to be dense while reconstructing the planes~\cite{wang2018learning} and the $\delta^2$ operator is defined according to the second order prior smoothness term $\mathcal{S}(\{j,k,l\})$ \cite{Woodford2008GlobalPriors}:
	\begin{equation}
	\mathcal{S}(\{j,k,l\},d_t^*)_x = \delta^2_x d_t^* = d_t^*(j) - 2* d_t^*(k) + d_t^*(l),
	\end{equation}
	with $\{j, k, l\}$ three neighboring pixels in the horizontal or vertical direction following the $x$-axis or $y$-axis orientation of the smoothing.

	The weighted combination $L_{\textrm{diff}}$ of these two terms then allows for a precise reconstruction of the non-reflective (diffuse) areas.

	\begin{equation}
	L_{\textrm{diff}} = \mu L_r + \lambda L_s,
	\end{equation}
	with $\lambda$ a scaling parameter set to $1e^{-3}$ and $\mu$ a binary mask defined in \cite{godard2019digging} taking occlusion and displacement of pixels along sequences into account.

Now, by drawing inspiration from and generalizing the contribution in \cite{berger2017depth}, it is possible to include a third term into this loss to penalize poor reconstruction of reflective areas. By definition, polarization is defined by the orientation of the electric field. Consequently, it is possible to estimate the electric field as a function of a normal derived from a plane.
	
	\begin{figure}[!t]
		\centering
		\vspace{0.2cm}
		\includegraphics[keepaspectratio,width=.8\linewidth]{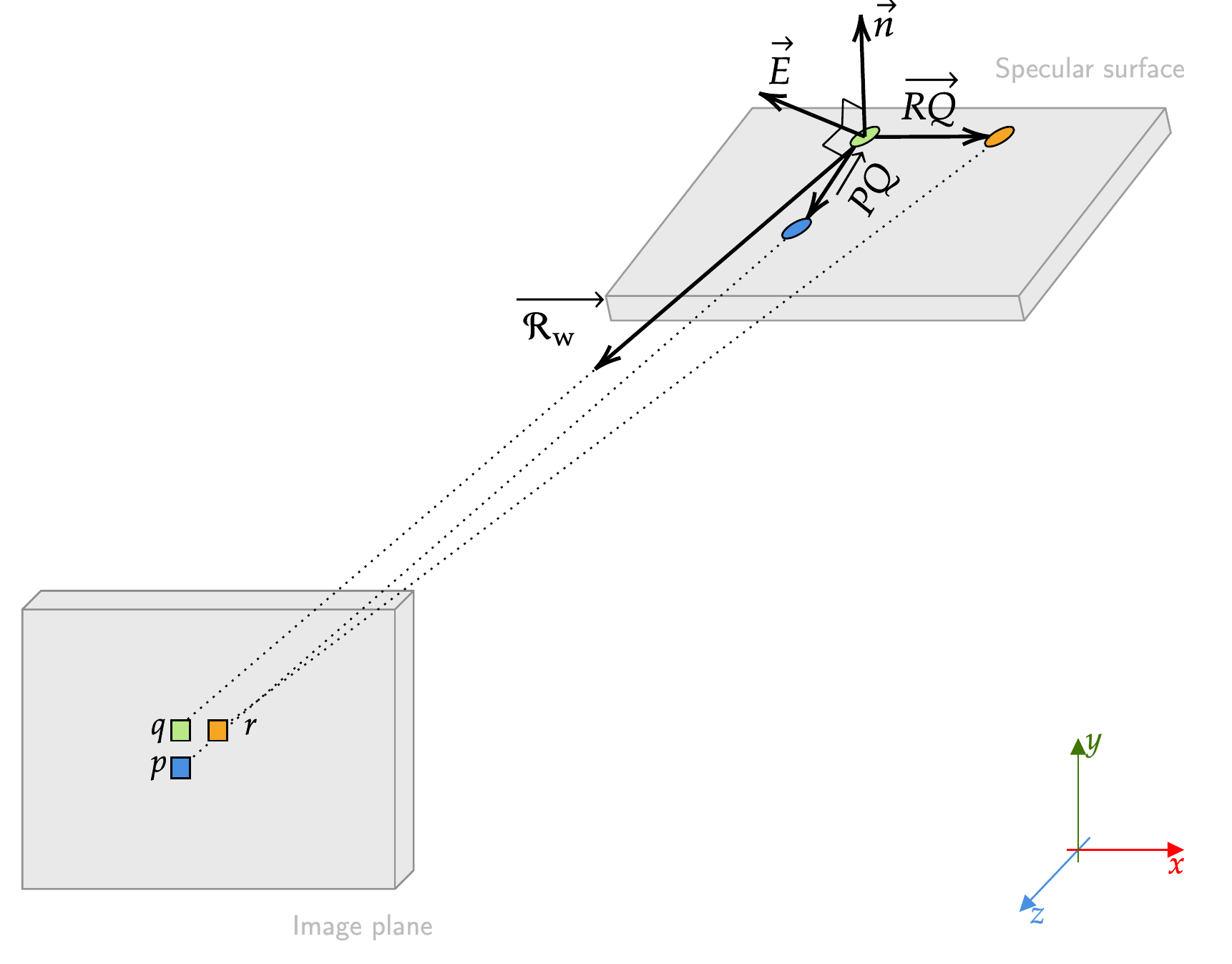}
		\caption{Illustration of the electric field estimation method.}
		\label{fig:efromp}
	\end{figure}
	
\begin{figure}[t]
		\centering
		\includegraphics[keepaspectratio,width=\linewidth]{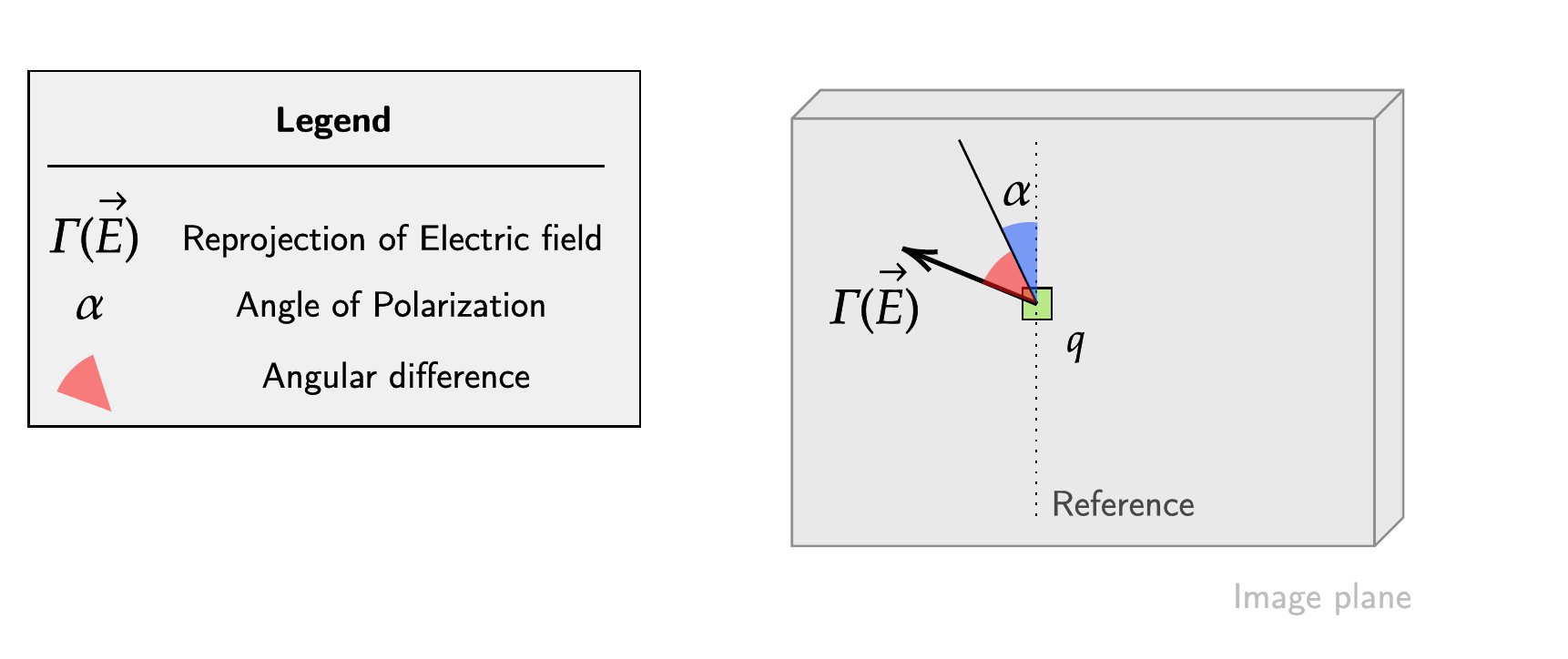}
		\caption{Angular difference visual representation. Here, the reference of the angle of polarization is vertical.}
		\label{fig:cpol}
	\end{figure}

	Let us consider three neighboring pixels $\{p, q, r\}$ in the layout presented in the Figure~\ref{fig:efromp}. This arrangement is organized such that it removes fronto-parallel planes related uncertainties. Then, the projection of these three adjacent pixels into 3D results in three points of a plane, respectively $\{ P, Q, R \}$. 
	The local normal $\overrightarrow{n}$ is obtained via the cross-product of the two vectors $\overrightarrow{PQ}$ and $\overrightarrow{RQ}$ linking the points $\{P,Q,R\}$. By definition, the electric field $\overrightarrow{E(Q)}$ is perpendicular to the plane defined by the the normal and the reflected wave when considering specular surfaces. Following the definition, $\overrightarrow{E(Q)}$ at 3D point $Q$ can be deduced from the cross product between the local normal and $\overrightarrow{\mathcal{R}_\mathrm{w}}$ at the point $Q$ as follows:
    
	\begin{equation}
	    \begin{split}
	\overrightarrow{E(Q)} = &\Bigl[ \Big(\Pi(p,D(p)) - \Pi(q, D(q))\Big) \times\\& \Big(\Pi(r,D(r)) - \Pi(q, D(q))\Big) \Bigr]
	\times \overrightarrow{\mathcal{R}_\mathrm{w}} ,
	\end{split}
	\end{equation}
	
	where $\Pi(x,D(x))$ is the 3D projection of pixel $x$ relative to the disparity $D(x)$. In an optimal context, the polarization angle and the electric field maintain the same orientation and by extension the same angle relative to the reference as shown in the Figure~\ref{fig:cpol}. Conversely, when a depth map is incorrectly estimated, then the estimated local normal is inconsistent and consequently is the deduced polarization angle. Accordingly, we can add a term $C_{\textrm{pol}}$ to the loss penalizing the deviation of the normal.

As shown in equation \ref{backpro} and in Figure~\ref{fig:cpol}, to evaluate the deviation, it is necessary to back project the direction of the electric field onto the image plane and compare it with the angle of polarization $\alpha$:
	
	\begin{equation}\label{backpro}
	C_{\textrm{pol}}(q) = \rho(q) ~ \Big|\tan\big[\tan^{-1}\Big(\Gamma(\overrightarrow{E(Q)})\Big) - \alpha(q)\big]\Big| ,
	\end{equation}
	where $\Gamma$ is the back projection operator onto the image plane. 
	Moreover, $\rho$ allows the scaling of the loss reinforcing the necessity for correlation between $\alpha$ and $\rho$.\\

	Since an angular differences is considered, and because this term will be combined with the reprojection term, the definition domains must be taken into account. The reprojection term clearly belongs to $[0,\infty[$ interval. To constrain the polarization term to the same interval, the absolute tangent is employed.
As a result, the polarimetric loss term becomes:
	\begin{equation}\label{LPol}
	L_{\textrm{pol}} = \frac{1}{N}\sum_{x \in \chi} C_{\textrm{pol}}(x),
	\end{equation}
	with $N$ the number of pixels $x$ in the set of reference image pixels $\chi$. Finally, the loss used to train the network is defined by: 
	
	\begin{equation}
	    \Lambda =  L_{\textrm{diff}} + ~\tau ~L_{\textrm{pol}},
	\end{equation}
	\noindent
	where $\tau$ is a binary mask derived from $\rho$ such that:
	\begin{equation}
	\tau(x) = \begin{cases} 1, & \mbox{ if } \rho(x) ~ \geq ~ 0.4\\
	0, & \mbox{ otherwise}\end{cases}.
	\end{equation}
	The polarimetric term $L_{\textrm{pol}}$ is taken into account only if the degree of polarization is relevant. Since, the relevance of both $\rho$ and $\alpha$ are correlated, this mask ensure for a legitimate electric field estimation.
The final loss $\Lambda$ is then just composed of reprojection error when the image area is unpolarized. Polarization components, when consistent, are taken into consideration and penalize the inaccurate reconstruction of specular surfaces. \\

\begin{figure}[t]
		\centering
		\vspace{.1cm}
		\includegraphics[keepaspectratio,width=\linewidth]{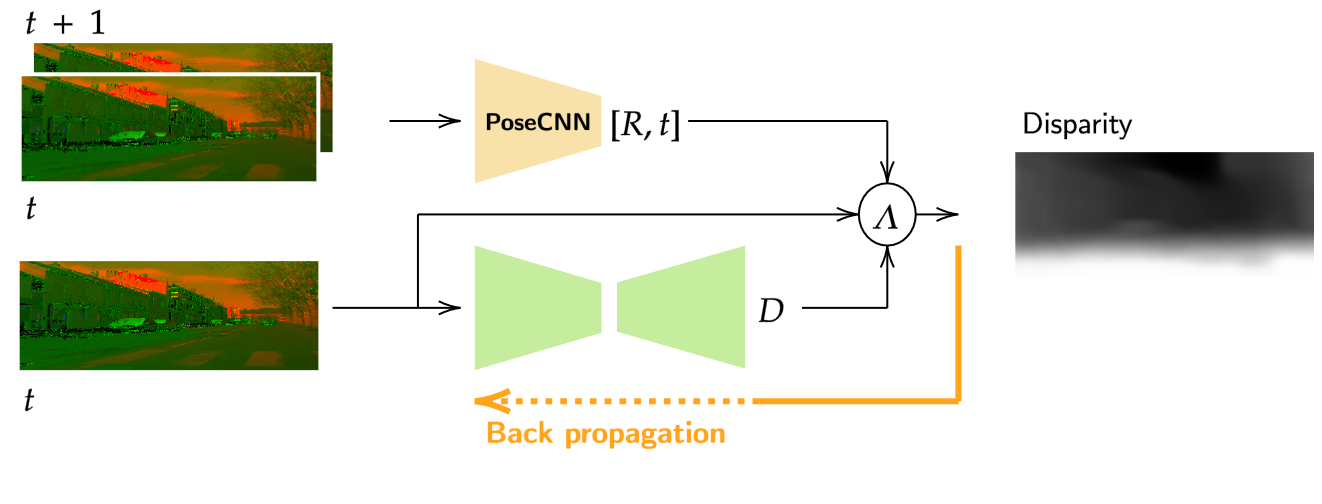}
		\caption{Illustration of the network as well as the loss calculation strategy and its back propagation. Drawing inspiration from \cite{godard2019digging}, the depth estimation network is a UNet with a ResNet50 layout.}
		\label{fig:net}
	\end{figure}

	\textbf{Final architecture.} 
	Following \cite{godard2019digging}, the network has an encoder-decoder architecture (a UNet with a ResNet 50 layout as shown in the Figure~\ref{fig:net}). It takes as input three-channel images obtained by concatenation of the intensity $\iota$, the polarization angle $\alpha$ and the degree of polarization $\rho$.
To overcome some inconsistencies related to the polarimetric modality and to consider exclusively areas with a minimum partial specularity, all values of lower than
0.4 are eliminated. This is justified by the fact that diffuse surfaces corresponding to low degree of polarization lead to a difference of $\pi/2$ between $\alpha$ and the electric field $E$.
When the disparity induced by the reprojection error is calculated, despite the accuracy of this calculation, the angular error will then tend towards $\pi/2$ leading the $L_{\textrm{pol}}$ function to tend towards infinity and thus causing exploding gradient problems.\\
Similarly, a perfect $\rho$ is physically unobservable which justifies an upper threshold. To combine a scale-factor effect and a regularization relative to physical property, the values are clipped to a maximum of 0.8.
	
	\section{Experiments}
	
	\subsection{Implementation details}
	
	\indent
	\textbf{Datasets.} The training dataset was acquired during both dry and rainy weather such that the experiments would highlight the capacity of polarimetric modality in diverse conditions.
All acquisitions were made with an affordable polarimetric camera, the Basler Ace aca2440-75um POL, consisting of a Sony IMX250MZR sensor delivering a resolution of 2448~x~2048 pixels.
The camera was mounted on board a driving car, recording a total of approximately 7,000 images per weather condition. The final training dataset is composed of 13,400 images.
As for the evaluation dataset, it is composed of a completely independent set of 25 images, acquired separately from another view under mixed meteorological conditions.\\
	
	\textbf{Ground truth. }Ground truth generation represents a critical point when it comes to addressing urban reconstruction problems.  Because of specularity, accurate depth evaluation is difficult since ground truth generation commonly rely on LiDAR sensors which are occasionally unreliable for measuring specular surfaces geometry due to reflection or transparency. Indeed, it would be a prerequisite to spray matte coating over all the specular surfaces of a complex urban scene wich is obviously unfeasible.
To overcome such difficulties, the reference disparity has been pre-calculated using SGBM \cite{hirschmuller2007stereo} and then refined by hand. It would have been possible to calculate the ground truth using a learning-based method. It should be considered the approach presented here is to improve deep learning methods since they typically fail on specular surfaces. Moreover, it is notable the vast majority of networks are trained on the same database which consists of images in favorable weather conditions. For these reasons, the choice of a refined SGBM eliminates learning biases while providing ground truth taking into account specular surfaces. This approach is unconventional but permits to conceive a global idea of the reliability of the results while allowing the computation of metrics. In addition, the disparity remains a relative value and therefore the impact of manual refinement is minor.\\

	\textbf{Network training. }The network was trained on a machine consisting of a Nvidia Titan Xp (12GB memory) GPU, 128GB of RAM and two CPUs accumulating a total of 24 physical cores. 
We use the following parameters for all the networks: a batch size of 12, a learning rate of 1e-3 and a maximum of 30 epochs.
For a fast training, the images were downsampled without any interpolation method to maintain the physical properties. 
Following this routine, training with polarimetric images takes approximately 17 hours compared to 12 hours when training with intensity images only. 
The forward pass inference time is around 0.45 second per image in pure CPU processing.\\
	
	\textbf{Evaluation.} We compared the results of our method P2D with the competitive state-of-the-art method described in \cite{godard2019digging}. Our P2D receives as input the polarization parameters by concatenation of the three channels $\{\iota, \alpha, \rho\}$. For the method in \cite{godard2019digging}, we evaluate two versions. One version, $G_{\textrm{RGB}}$, using only intensity images and trained with the weights provided by the authors without fine-tuning. And, another version, $G_{\textrm{I}}$, trained in an end-to-end manner so that the network parameters are adapted to the intensity images at hand.\\

	\textbf{Metrics. }The calculated metrics shown in the table \ref{resmet} represents popular assessments within the reconstruction community that have been proposed by Eigen et al. \cite{eigen2014depth}. They provide an unbiased and comprehensive measure of results. In particular, the $\delta$ values are calculated on the prediction/ground truth ratio and highlight an intrinsic precision of the reconstruction.\\
	
	At last, the sky reconstruction accuracy $R_s$ is calculated as follows:
    \begin{equation}\label{reconSky}
        R_s = 1 - \Big(\frac{\hat{y}_s}{y_s}\Big),
    \end{equation}
    where $y_s$ is the sum of the binary masked pixels considered as sky in the ground truth and $\hat{y}_s$ the corresponding area in the prediction. This calculation is performed on the disparity, and one order of magnitude error deviation is considered acceptable.
It focuses on the ability of the network to accurately estimate the sky and not propagate an erroneous evaluation in such areas. It is noteworthy this kind of precision is usually neglected since the reconstruction precision of these areas is removed from the frequent metrics. Ordinarily, sky zones are filtered out of the metrics beforehand. In this contribution, these areas are also neglected while calculating Eigen et al.~\cite{eigen2014depth} metrics.
	
\renewcommand{\arraystretch}{1.1}
\begin{table*}[!t]
	\centering
	\caption{Quantitative comparative results. For each network several metrics are computed neglecting the sky areas. In addition, we propose three different evaluations: on the \textit{Raw} images at the output of the network, on the \textit{Cropped} images to eliminate inconsistencies in the polarimetric network, and on the \textit{Specular} areas only.
	G$_{\textrm{RGB}}$ corresponds to the network presented in \cite{godard2019digging} without fine-tuning, and G$_{\textrm{I}}$ corresponds to the same network with fine-tuning.
P2D corresponds to our method.}\label{resmet}
 \center
	
		\begin{tabular}{c||c||cccc||ccc}
			\hline
			\small\textbf{Type}  & \small{\textbf{Network}}  & \small{Abs\_Rel}  & \small{Sq\_Rel}   & \small{RMSE}  & \small{RMSE\_log} & \small$\delta > 1.25$    & \small$\delta > 1.25^2$    & \small$\delta > 1.25^3$    \\ \hline \hline
			\parbox[t]{5mm}{\rotatebox[origin=c]{90}{\textit{Raw}}} & \begin{tabular}[c]{@{}c@{}}G$_{\textrm{RGB}}$\\ G$_{\textrm{I}}$\\ P2D\end{tabular} & \begin{tabular}[c]{@{}c@{}}0.471\\ 0.482\\ \textbf{0.322}\end{tabular} & \begin{tabular}[c]{@{}c@{}}10.809\\ 9.144\\ \textbf{4.504}\end{tabular} & \begin{tabular}[c]{@{}c@{}}25.161\\ 22.332\\ \textbf{20.651}\end{tabular} & \begin{tabular}[c]{@{}c@{}}0.680\\ 0.617\\ \textbf{0.484}\end{tabular} & \begin{tabular}[c]{@{}c@{}}0.485\\ 0.431\\ \textbf{0.537}\end{tabular} & \begin{tabular}[c]{@{}c@{}}0.707\\ 0.695\\ \textbf{0.801}\end{tabular} & \begin{tabular}[c]{@{}c@{}}0.804\\ 0.838\\ \textbf{0.896}\end{tabular} \\ \hline \hline
			\parbox[t]{5mm}{\rotatebox[origin=c]{90}{\textit{Cropped}}} & \begin{tabular}[c]{@{}c@{}}G$_{\textrm{RGB}}$\\ G$_{\textrm{I}}$\\ P2D\end{tabular} & \begin{tabular}[c]{@{}c@{}}0.533\\ 0.415\\ \textbf{0.245}\end{tabular} & \begin{tabular}[c]{@{}c@{}}14.050\\ 11.247\\ \textbf{5.650}\end{tabular} & \begin{tabular}[c]{@{}c@{}}29.312\\ 25.899\\ \textbf{24.009}\end{tabular} & \begin{tabular}[c]{@{}c@{}}0.780\\ 0.678\\ \textbf{0.531}\end{tabular} & \begin{tabular}[c]{@{}c@{}}0.449\\ 0.467\\ \textbf{0.604}\end{tabular} & \begin{tabular}[c]{@{}c@{}}0.658\\ 0.729\\ \textbf{0.825}\end{tabular} & \begin{tabular}[c]{@{}c@{}}0.771\\ 0.850\\ \textbf{0.910}\end{tabular} \\ \hline \hline
			\parbox[t]{5mm}{\rotatebox[origin=c]{90}{\textit{Specular}}} & \begin{tabular}[c]{@{}c@{}}G$_{\textrm{RGB}}$\\ G$_{\textrm{I}}$\\ P2D\end{tabular} & \begin{tabular}[c]{@{}c@{}}0.341\\ 0.208\\ \textbf{0.147}\end{tabular} & \begin{tabular}[c]{@{}c@{}}8.249\\ 2.248\\ \textbf{1.583}\end{tabular} & \begin{tabular}[c]{@{}c@{}}7.236\\ 5.491\\ \textbf{4.898}\end{tabular} & \begin{tabular}[c]{@{}c@{}}0.306\\ 0.233\\ \textbf{0.166}\end{tabular} & \begin{tabular}[c]{@{}c@{}}0.666\\ 0.639\\ \textbf{0.796}\end{tabular} & \begin{tabular}[c]{@{}c@{}}0.808\\ 0.877\\ \textbf{0.921}\end{tabular} & \begin{tabular}[c]{@{}c@{}}0.896\\ 0.952\\ \textbf{0.973}\end{tabular} \\ \hline
		\end{tabular}
	
\end{table*}
	\subsection{Results and discussion} 
\begin{figure*}[!ht]
		\centering
		\includegraphics[keepaspectratio,width=.8\linewidth]{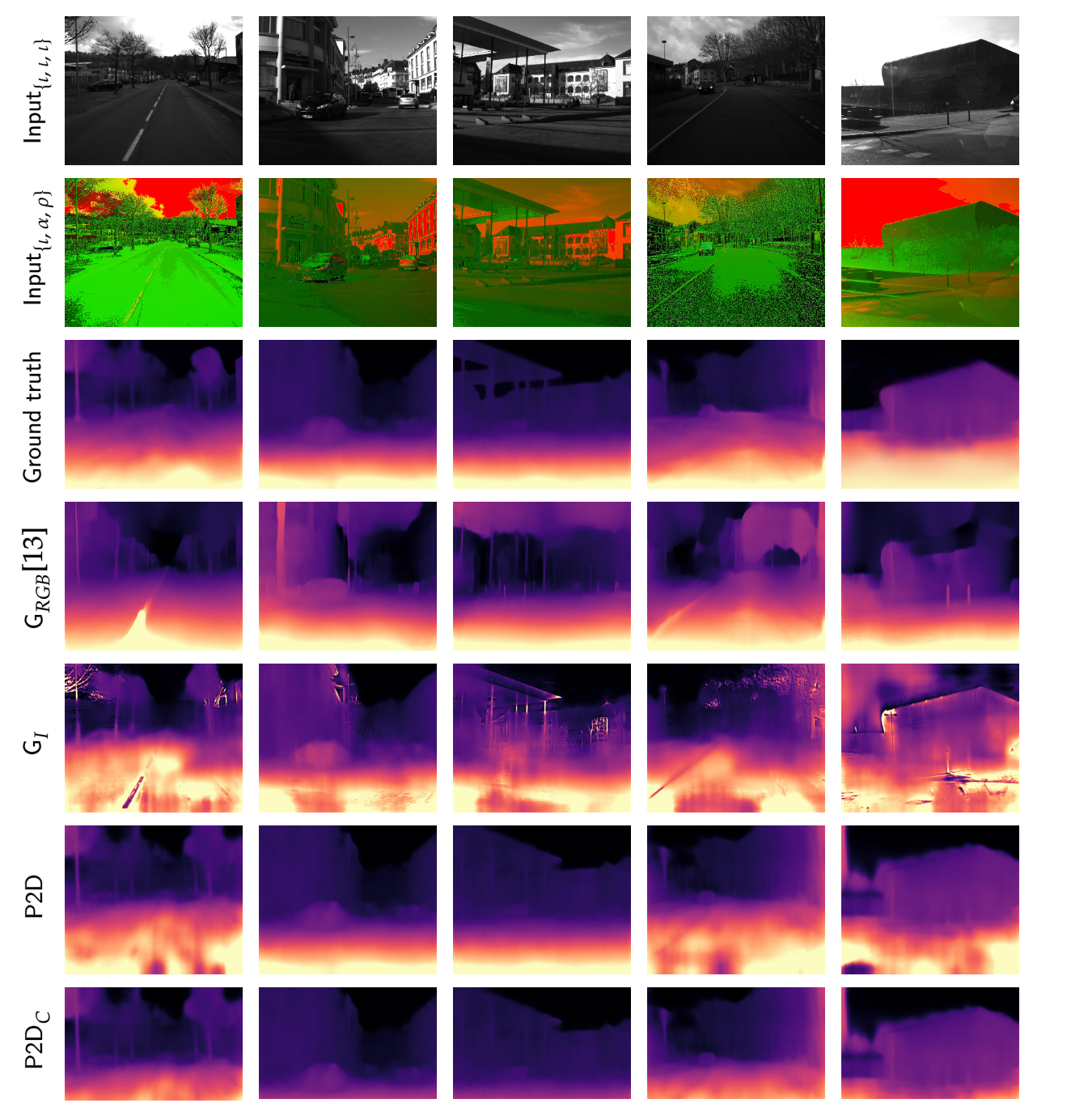}
		\caption{Illustration of results on five independent road scenes in mixed weather conditions. From top to bottom, the inputs to the two networks (scalar or polarimetric), the ground truth depth map, then the results of the three different networks: G$_{\textrm{RGB}}$ corresponds to the network presented in \cite{godard2019digging} without fine-tuning, and G$_{\textrm{I}}$ corresponds to the same network with fine-tuning. P2D corresponds to our method.
        The last row shows the crop version of the results from P2D to eliminate inconsistencies due to the modality and aberrations because of the camera position. Columns one and four correspond to acquisitions in light rainy weather, hence, the different road behaviour in polarimetric space. The other images are acquired under normal conditions.\vspace{.5cm}}
		\label{fig:res}
		
	\end{figure*}

	Table~\ref{resmet} and Figure~\ref{fig:res} allow for a quantitative and qualitative evaluation of the results. 
Analyzing the images in Figure~\ref{fig:res}, we can observe various responses of the networks.
First, using the method in \cite{godard2019digging} with raw images (G$_{\textrm{RGB}}$), the results seem satisfactory at first glance. However, some characteristics of the images are altered. For example, specular areas, car windshields or bus stops are incorrectly detected. To be specific, car windshields are over-segmented into several parts rather than being detected as unique planar surface.
In addition, the distance to reflective road lines is often under-estimated and farthest objects are ignored.
However, as the weights of the network are not fine-tuned, its features representations have been learned exclusively from textures characteristics which limit the performance of the method in specular or reflective areas. 
Nevertheless, the reconstruction is close enough to the ground truth which also shows the robustness of this approach and reinforces the initial idea of using it as a baseline method.

	When the network is trained end-to-end with polarimetric intensity images, the global impact of polarization is reduced but brings many significant constraints. Despite an accurate estimation of some areas like planar surfaces on cars and long distance objects (see row G$_I$ in Figure~\ref{fig:res}), others areas are subject to some aberrations mainly on the reflective lines and polarized contours. This behaviour produces a direct impact on the network estimations.
Hence, the addition of polarization-specific terms is necessary to improve the estimation of polarized areas.

When employing all the polarimetric information ($\iota$, $\rho$ and $\alpha$) in our P2D network, we can observe a accurater estimation of specular areas as well as sufficient reconstruction of diffuse areas.\\
We can however see that the results at limited distances, as shown in the P2D row of Figure~\ref{fig:res}, are occasionally incorrect.
This is due to the fact polarimetric information varies according to the light and its reflection angle. Therefore, the position of the camera is primarily responsible for these erroneous estimates.
Indeed, as explained earlier in the Datasets section, the images of the evaluation subsets were acquired with different camera poses. Consequently, the information from the images differs from the training set case leading the network to fail in estimating depth values at close distances. 
To have an estimation in favorable conditions, the choice of cropping the lower quarter of the images for a second evaluation is proposed. Note, this lower part corresponds to closer distances. Both the estimates and the ground truth depth maps are cropped. 
These results are shown in the P2D$_\textrm{C}$ line of Figure~\ref{fig:res} as well as the \textit{Cropped} part of the Table~\ref{resmet}. We can observe better performances especially when comparing errors with the $\delta$ metric. The most considerable improvements are obtained when looking at both $\delta > 1.25$ and $\delta > 1.25^2$ showing a respective 16\% and 17\% improvement compared to the $G_{\textrm{RGB}}$ network.\\

Additionally, we perform an evaluation considering only the specular areas of the scenes. This is achieved using a rule-based naive system filtering polarization degrees higher than 0.4, hence keeping areas which are highly specular. 
The results shown in the last part of Table~\ref{resmet}, exhibit improved performances for all the evaluated networks since the assessed pixel space is reduced. However, the largest improvement is obtained with our P2D method. Specifically, our method achieves 92\% for $\delta > 1.25^2$ ratio, compared to 80\% obtained by the state-of-the-art method.
Consequently, we can see the polarimetric modality is beneficial for the reconstruction of urban scenes with many specular surfaces. 
Ultimately, to highlight the depth map reliability, sky reconstruction accuracy (eq.~\ref{reconSky}) has been computed in Table~\ref{Reconsky}.\\

	\begin{center}
\begin{table}[H]
\centering
\caption{Quantitative comparison of sky reconstruction accuracy.}
\label{Reconsky}
\begin{tabular}{c||ccc}
\hline
Network & $G_{RGB}$ & $G_{I}$ & P2D   \\ \hline \hline
$R_s$      & 0.055     & 0.388   & \textbf{0.532} \\ \hline
\end{tabular}
\end{table}
\end{center}
This specific metric has been computed since many evaluation metrics neglect such aspect which, however, could be informative, especially if one uses such an algorithm for navigation. This ratio reveals P2D's ability to reconstruct slightly more than half of the sky correctly. It permits to demonstrate polarimetry to be favorable also for such estimation.

	\section{Conclusion}
	
	In recent years, the cost of polarimetric cameras has been on a decreasing trend. This allows the use of a more affordable alternative type of information and thus an opening to data directly related to the physics of the scenes. 
We have proposed a method to include polarization cues in a cost function applied to monocular depth estimation. With this polarimetry adapted loss, we can directly consider geometric constraints specific to the scene.

We next showed the proposed approach was remarkably effective for depth estimation in urban environment. Moreover, unlike previous methods, polarization allows being insensitive to specular surfaces. This is a significant advantage since traditional algorithms have difficulties to cope with such areas.

It is explicit that a fusion method using an RGB and polarimetric pair would allow an accurate and stable reconstruction in any condition. Despite this observation, one of the considerable challenges of this approach remains the multimodal/multifocal alignment.
Recent sensors, like the Sony Pregius IMX250MYR-C, would potentially allow combining the strengths of each modality by merging the diverse types of information.
It would be beneficial to take advantage of this modern technology to propose an accurate reconstruction of both specular and diffuse areas. Ultimately, it is noticeable that RGB aligned images would allow a completely fair and unbiased comparison for the methods.

	\section*{Acknowledgments}

This work was supported by the French National Research Agency through ANR ICUB (ANR-17-CE22-0011). We gratefully acknowledge the support of NVIDIA Corporation with the donation of GPUs used for this research.   
	
	\bibliographystyle{IEEEtran}
	\bibliography{egbib,refs}
	
\end{document}